\def\year{2019}\relax
\documentclass[letterpaper]{article} 
\usepackage{aaai19}  
\usepackage{times}  
\usepackage{helvet} 
\usepackage{courier}  
\usepackage[hyphens]{url}  
\usepackage{graphicx} 
\def\UrlFont{\rm}  
\usepackage{graphicx}  
\usepackage{color}
\usepackage{amsmath}
\usepackage{subcaption}
\usepackage{tikz}
\usetikzlibrary{shapes,arrows,positioning}
\usepackage{multirow}
\usepackage{array}

\newcommand{\citet}[1]{\citeauthor{#1}~\shortcite{#1}}
\newcommand{\citep}{\cite}

\newcommand{\hide}[1]{} 

\begin{document}
%
\title{Adapting Sequence to Sequence models for Text Normalization in Social Media}
\author{
Ismini Lourentzou,
Kabir Manghnani,
ChengXiang Zhai \\
University of Illinois at Urbana-Champaign, Urbana, IL, USA \\
\{lourent2,kabirm2,czhai\}@illinois.edu
}
\maketitle

\begin{abstract}
\hide{The advancement of NLP research has relied on the availability of academic annotated datasets from clean formal textual sources, such as books and news articles.} Social media offer an abundant source of valuable raw data, however informal writing can quickly become a bottleneck for many natural language processing (NLP) tasks. Off-the-shelf tools are usually trained on formal text and cannot explicitly handle noise found in short online posts. Moreover, the variety of frequently occurring linguistic variations presents several challenges, even for humans who might not be able to comprehend the meaning of such posts, especially when they contain slang and abbreviations. Text Normalization aims to transform  online user-generated text to a canonical form. Current text normalization systems rely on string or phonetic similarity and classification models that work on a local fashion. We argue that processing contextual information is crucial for this task and introduce a social media text normalization hybrid word-character attention-based encoder-decoder model that can serve as a pre-processing step for NLP applications to adapt to noisy text in social media. Our character-based component is trained on synthetic adversarial examples that are designed to capture errors commonly found in online user-generated text. Experiments show that our model surpasses neural architectures designed for text normalization and achieves comparable performance with state-of-the-art related work.
\end{abstract}

\section{Introduction}
Most text data in the world today is user-generated and online. Vast quantities of online blogs and forums, social media posts, customer reviews, and other textual sources are necessary input of useful information for algorithms that understand user intent and preferences, predict trends or recommend items for purchase in targeted advertising. 
However, social media usually deviates from standard language usage, with high percentages of non-standard words, such as abbreviations, phonetic substitutions, hashtags, acronyms, internet slang, emoticons, grammatical and spelling errors.

Such non-standard words cause problems for both users and text mining applications. Users that are not familiar with domain-specific or peculiar language usage, e.g. acronyms found in Twitter messages, may experience problems in understanding the expressed content. Additionally, due to high out-of-vocabulary word rates, NLP approaches struggle with the noisy and informal nature of social media written language. Natural language follows a Zipfian distribution where most words are rare. Learning ``long tail" word representations requires enormous amounts of data \citep{bahdanau2017learning}.
Recent work showcases the negative impact of noisy text on several NLP tasks and the improvement that text normalization can bring in  part-of-speech tagging \citep{han2013lexical}, parsing \citep{zhang2013adaptive}, and machine translation \citep{hassan2013social} tasks. Special pre-processing of informal text is therefore necessary to help users understand content more easily and facilitate NLP algorithms. The task of transforming noisy or informal text to a more standard representation is called \textbf{\textit{Text Normalization}}.

Normalizing text is challenging and involves a trade-off between high recall, i.e. maximizing the number of corrections and high precision, i.e. minimize the number of incorrect normalizations. In several cases, the task is framed as mapping  an out-of-vocabulary (OOV) non-standard word to an in-vocabulary (IV) standard one that preserves the meaning of the sentence. Additionally, text normalization can include modifications that are beyond the framework described above, for example replacing, removing or adding word tokens or punctuation and capitalize or lowercase text. Word mappings might not be unique, i.e. an OOV word can be transformed to more than one IV word, based on context. Due to the dynamic nature of social media text, many words (e.g. named entities) are considered OOV but do not need normalization or there is no appropriate IV word for them.

Although text normalization may appear to be similar to the task of spelling error correction, it is actually much more difficult to handle noisy social media text.  Spelling correction focuses on word errors that can usually be handled with edit distance metrics. Additionally, grammatical error correction, which incorporates local spelling errors with global grammatical errors, e.g. preposition or verb usage mistakes, deals with replacing or adding omitted words, which are often caused \textit{unintentionally} by non-native writers. Such errors can be partially identified with syntactic knowledge, e.g. semantic parsing, while it is rather unlikely that text normalization systems will benefit from such linguistic sources \citep{baldwin2015shared}. Due to the new challenges in text normalization, it generally requires new approaches that go beyond the traditional spelling error correction methods. 

The non-standard forms found in user-generated context can be mostly summarized into several categories:
\begin{enumerate}
    \item \textbf{Misspellings}, e.g. ``defenitely" $\rightarrow$ ``definitely"
    \item \textbf{Phonetic substitution} of characters with numbers or letters, e.g. ``2morrow" $\rightarrow$ ``tomorrow", ``rt" $\rightarrow$ ``retweet"
    \item \textbf{Shortening} of words, e.g. ``convo" $\rightarrow$ ``conversation" 
    \item \textbf{Acronyms}, e.g. ``idk" $\rightarrow$ ``i don't know" that can also include standard words usually used as acronyms, e.g. ``goat" $\rightarrow$ ``greatest of all time"
    \item \textbf{Slang}, i.e. metaphoric usage of standard words, e.g. ``low key", ``woke" or ``broccoli" 
    \item \textbf{Emphasis} given to a certain word, either by capitalization, e.g. ``YEAH THIS IS SO COOL" or by vowel elongation e.g. ``cooooool"  $\rightarrow$ ``cool"
    \item \textbf{Punctuation} deleted or misplaced, e.g. ``doesnt"  $\rightarrow$ ``doesn't", ``do'nt"  $\rightarrow$ ``don't" or intentionally using punctuation instead of letters, e.g. ``f@ck"
\end{enumerate}

Early text normalization systems consider a pipeline of statistical language models, dependency parsing, string similarity, spell-checking and slang dictionaries \citep{liu2011insertion2011a,han2011lexical,han2013lexical}. However, the high-dimensional action space of language (arbitrary word sequences constructed from a vocabulary) makes unsupervised learning inefficient. 
Additionally, unsupervised text normalization methods often tune hyper-parameters based on annotated (supervised) data, thus are not fully unsupervised.
Considering the rapid changes of language in online content, with many emerging words appearing daily, lexicon-based approaches are not able to handle social media text properly. String similarity, such as edit distance, does not work on non-standard words where the number of edits is large, for example abbreviations. In order to achieve better pre-processing performance, we need to develop methods that are specifically designed for the problem at hand. 

\begin{figure}
    \centering
    \includegraphics[width=\columnwidth]{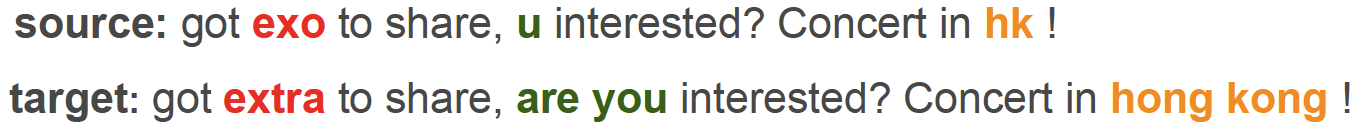}
    \caption{Example of source (unnormalized) tweet and target (normalized) pair where context helps in correcting ambiguous terms. The word ``exo" needs to be transformed to ``extra", while the word ``concert" provides the required context to understand that ``exo" refers to an extra ticket.}
    \label{fig:example}
\end{figure}

Recent work relies on candidate generation and ranking (see section ``Related work" for a thorough review), with two major deficiencies: Current approaches have mostly ignored the contextual information present in a sentence that can be potentially very useful. More specifically, in most cases the features extracted or the models developed are \textit{limited to a specific context window}, e.g. \citet{min2015ncsu_sas_wookhee} work with character-ngrams, while \citet{jin2015ncsu} relies on features that depend on previous, current and next tokens of a candidate term. This requires additional human effort to decide the appropriate ngram order and design features. More importantly, it restricts the system in a way that prevents longer contextual dependencies to be leveraged (see Figure \ref{fig:example} for an example).  The second limitation is that correcting complex normalization mappings are harder to tackle and methods that rely on candidate generator functions by definition limit their approach to specific types of errors. For example, it would be difficult for such methods to handle multiple normalization errors at once, e.g. spelling errors on an acronym or a slang term, and combining candidate generator functions results in a combinatorial problem. 

Drawing inspiration from neural machine translation \citep{bahdanau2014neural,luong2015effective}, 
we propose to address both limitations by using end-to-end neural network models, particularly sequence-to-sequence (Seq2Seq) models. Specifically, Seq2Seq models can encode the entire sequence data with hidden neurons that would naturally capture any useful context information in a sentence to improve text normalization performance. Such models can handle complex normalizations without the need of language-specific tools, given enough  training data on any language (i.e. pairs of unnormalized and normalized sentences). 
We propose and compare several variants of Seq2Seq models for solving the problem of text normalization. 
The first is a straightforward application of the basic word-level Seq2Seq models to text normalization. 
However, such an approach faces a major challenge of high percentage of OOV tokens. 
In contrast, character-level Seq2Seq models do not operate on a limited vocabulary but are much slower to train and recent work has shown that character-based sequence-to-sequence models are not robust on noisy data \citep{belinkov2017synthetic}. We propose a novel hybrid end-to-end model that takes into account contextual information as well as addresses the OOV problem in a more robust way. Specifically, our model is based on a recurrent neural encoder-decoder architecture that reads the informal text sequences, transforms them in a continuous-space representation that is passed on the decoder to generate the target normalized sequence. To capture local spelling errors and morphological variations of OOV words, we correct unknown words with a character-level encoder-decoder trained on synthetic \textit{adversarial examples} that capture common errors in online user-generated text. Our method obtains open vocabulary coverage while maintaining the lower training time of word-based models, when compared with character-level sequence-to-sequence architectures. 

Our contribution is two-fold: 
\begin{enumerate}
    \item We explore variations of encoder-decoder architectures as well as adversarial training for tackling the task of normalizing social media text.
    \item We propose a novel hybrid  neural network architecture specifically designed for normalizing OOV tokens. Coupled with adversarial training, our model allows for an open set of corrections while seamlessly incorporates context and long-term dependencies. 
    Through carefully designed experimentation, we show that the proposed hybrid model outperforms both word and character based standard Seq2Seq architectures.
\end{enumerate}

Our source code and model files are publicly available \footnote{\url{https://github.com/Isminoula/TextNormSeq2Seq}}.

\section{Related Work}\label{sec:soa}
We briefly discuss related work on text normalization. The normalization problem was originally framed as standardizing numbers, dates and acronyms in formal text \citep{sproat2001normalization} but the definition was later broaden to transform social media informal text into canonical forms that NLP tools were usually trained on \citep{sproat2001normalization}. Research on this problem adopts several paradigms, from spell checking \citep{choudhury2007investigation}, machine translation 
\citep{aw2006phrase,ling2013paraphrasing} and speech recognition \citep{kobus2008normalizing}.

Early unsupervised methods include probabilistic models \citep{cook2009unsupervised}, string edit distance metrics \citep{contractor2010unsupervised}, construction of normalization dictionaries \citep{han2012automatically,gouws2011unsupervised} or extracting training data from search results with carefully designed queries \citep{liu2011insertion2011a}. 
Another line of work is lexicon-based methods and classification models. \citet{han2011lexical} train a classifier that detects non-standard words and then generate candidates based on morphological and phonemic similarity metrics plus a normalization lexicon.
Other approaches include word association graphs \citep{sonmez2014graph}, random walks \citep{hassan2013social}, statistical \citep{beaufort2010hybrid,zhang2013adaptive} and log-linear models \citep{yang2013log} and ranking candidates with language models \citep{han2013lexical}. These methods are quite limited as they rely primarily on string and phonetic similarity for identifying lexical variations.

Recently, there is a growing trend on applying Deep Learning in a variety of areas, such as Computer Vision or NLP. Such models offer flexibility, can learn representations and tasks jointly and have produced state-of-the-art for several applications, e.g. object recognition, sentiment analysis or machine translation. The representational power of neural models can potentially allow learning of complicated text transformations, automatically handle language drift and work with heterogeneous large streams of user-generated text. We briefly describe state-of-the-art models and models that either leverage deep learning or contain a component that is trained with neural networks. \citet{chrupala2014normalizing} leverages unlabeled data by incorporating character embeddings as features in a model that learns to perform edit operations. \citet{sridhar2015unsupervised} used distributed representations of words trained on a large Twitter dataset to extract normalization lexicons based on contextual similarity. Similarly, \citet{ansari2017improving} leverage word embeddings, as well as string and phonetic similarity to match OOV words to IV words (1:1 mapping), however their method does not take into consideration contextual information and thus cannot properly handle cases where multiple canonical forms of a non-standard word are available. In contrast to this line of work, we do not rely on pretrained embeddings; we learn both word and character representations, in addition to the text normalization model in an end-to-end fashion.

\citet{baldwin2015shared} present a text normalization task on English tweets as part of the 2015 ACL-IJCNLP Workshop on Noisy User-generated Text (W-NUT). Two categories were introduced based on whether external resources and public tools were used (unconstrained systems) or not (constrained systems). Deep learning methods and lexicon-augmented conditional random fields (CRFs) achieved the best results, while the performance of unconstrained systems was inferior, suggesting that the underlying model produces a bigger impact on the final performance, compared with the usage of additional data or resources. The models of all participating teams are described in \citep{baldwin2015shared}. We should note that no team explored deep sequence-to-sequence models or adversarial training.

\citet{jin2015ncsu} achieved the best performance in the W-NUT task, with a method that generates candidates based on the training data. A binary random forest classifier is trained to predict whether a candidate is the correct canonical form for a token found in a tweet instance. Their feature set used during training includes string similarity, POS and statistics such as support and confidence. The final canonical form selected is the one that achieves the highest confidence score.
\citet{van2017monoise} extends this work by leveraging additional external resources of clean and informal text data.
\citet{min2015ncsu_sas_wookhee} combine a lexicon extracted from the training data with a recurrent neural model that performs edit operations trained on character trigrams. \citet{leeman2015ncsu_sas_sam} use a neural network classifier that predicts whether a word needs normalization and a secondary model that takes as input a word and outputs its canonical form. Framing the task of text normalization as classification in addition to relying on candidate generator functions limits the types of transformations that can be tackled, as candidate generation relies heavily on human engineering effort and existing methods for creating candidates cannot handle multiple complex normalization errors at once.

The aforementioned methods do not take into account the \textit{\textbf{full context}} in which a token appears. It is quite reasonable then for one to wonder whether Seq2Seq models would be appropriate for the task and in which ways the input or architecture should be adjusted in order to reach comparable performance given the limited training data available. To this end, we explore encoder-decoder models and study how crucial context is for the normalization of user-generated text. Finally, we design a novel hybrid Seq2Seq model that is trained on synthetic adversarial examples of noisy social media text.

\subsection{Sequence-to-Sequence Learning}
Encoder-decoder architectures \citep{sutskever2014sequence,cho2014learning} have been applied in a wide variety of natural language tasks, such as machine translation \citep{wu2016google}, dialogue generation \citep{vinyals2015neural}, summarization \citep{nallapati2016abstractive}, question answering \citep{yin2015neural}. Several extensions of the Seq2Seq models have been proposed with mechanisms such as attention \citep{bahdanau2014neural}, copying \citep{see2017get} and coverage \citep{tu2016modeling}.

In most cases only the most frequent words are kept, creating a fixed-sized vocabulary, with OOV words mapped to a common UNK token. Consequently, the performance is affected by the limited vocabulary. Recent work propose methods to mitigate this problem, by treating text as a sequence of characters \citep{lee2017fully}, inventing new word segmentation methods  \citep{sennrich2015neural,bojanowski2017enriching} or hybrid word-level models with an additional character-level model to handle problematic cases \citep{luong2016achieving,ji2017nested}.  While character-based models outperform models based on subword units, their extremely high computational cost and inability to handle long-distance dependencies makes them unappealing in practice. Moreover, as hybrid models only use the secondary character model for problematic cases, such as unknown words, they rely on large training datasets, making them inappropriate for domains with limited annotated data and frequent word variations. Our work lies on the hybrid models category but builds upon the properties of text normalization to adjust the character-based model training. 

\section{Text Normalization}\label{basic}
\begin{figure*}
    \centering
    \includegraphics[width=0.85\textwidth]{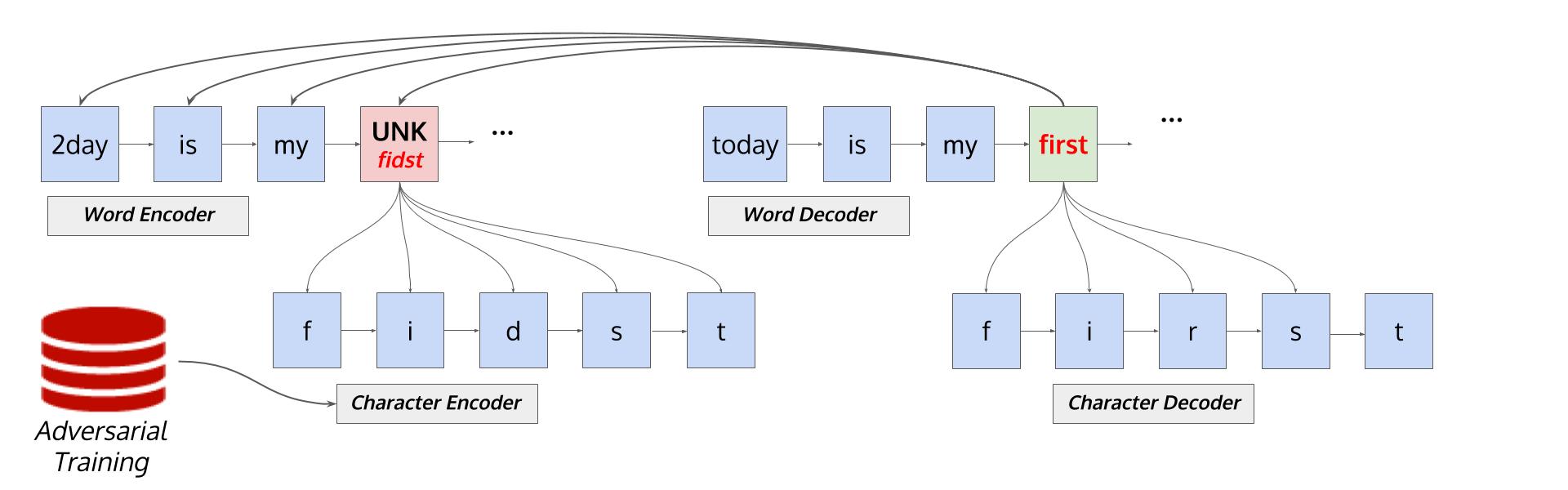}
    \caption{Our Hybrid Sequence-to-Sequence (HS2S) architecture that consists of two nested encoder-decoder architectures, one trained on word-level information and a character-based trained on synthetically generated adversarial examples. The primary model (word encoder-decoder) is trained on sequences of words. When an unknown symbol is encountered, such as token "fidst" (red box) in our example, we leverage a secondary character-level Seq2Seq model (character encoder-decoder) that is trained on a large pool of synthetic adversarial training examples of words to correctly normalize, e.g. token "first" (green box)}
    \label{fig:model}
\end{figure*}

Our architecture consists of two encoder-decoder models, primarily a word-based Seq2Seq model, while for transforming words not found in the word-level model's vocabulary, we either backtrack to a secondary character-based Seq2Seq model when its confidence is high or copy the source token (Figure \ref{fig:model}). For completeness, we briefly describe encoder-decoder neural models.

\subsection{Word-level sequence-to-sequence model}
Given an unnormalized text represented as an input sequence of words $\vec{x} = [x_1, \dots, x_T]$ with length $T$, we consider generating another output sequence of words $\vec{y} = [y_1, \dots, y_L]$ with length $L$ that has the same meaning as $\vec{x}$.  
The task is defined as a sequence-to-sequence learning problem which aims to learn the mapping from one sequence to another. Specifically, the architecture is built based on the encoder-decoder framework \citep{cho2014learning,sutskever2014sequence}, both of which are parameterized by attention-based recurrent neural networks (RNN). 

The encoder module reads the input sequence $\vec{x}$ and transforms it to a corresponding context-specific sequence of hidden states $\vec{h} = [h_1, \dots, h_T]$. In bi-directional models, two encoders are used; one reading the text in forward mode and another one reading text backwards. The final hidden state at time t  is the concatenation of the two encoder modules $\vec{h}_t = [g_{f}(x_t,h_{t-1}); g_{b}(x_t,h_{t+1})]$ where $g_{f}$ and $g_{b}$ denote the forward and backward encoder units, respectively. Similarly, the decoder defines a sequence of hidden states $\vec{s}_j = g_s(s_{j-1}, y_{j-1},  c_j)$ that is conditioned on the previous word $y_{j-1}$ and decoder state $s_{j-1}$, as well as the context vector $c_j$, computed as a weighted sum of encoder hidden states based on the attention mechanism \citep{bahdanau2014neural}:
$$c_j = \sum_{i=k}^{|T|} \alpha_{jk}h_{k}$$
where $\alpha_{jk} = \text{Softmax}(f(s_{j-1}, h_k))$ and $f(s_{j-1}, h_k) = s^{T}_{j-1}Wh_k$ is the \textit{general} content-based function described in \citet{luong2015effective}. Then, each target word is predicted by a Softmax classifier  $y_j \sim  p(y_j|y_{<j},\vec{x}) = Softmax(\psi(s_j))$, where $\psi$ is an affine transformation function that maps the decoder state to a vocabulary-sized vector.

Given training data $\mathcal{D}$, Seq2Seq model is trained by maximizing the log-likelihood:
$$ L(\theta) = - \sum_{(\vec{x},\vec{y}) \in \mathcal{D}} \sum_{j=1}^{|L|} \log p_{\theta}(y_j|y_{<j},\vec{x})$$
Note that during training a wrong prediction will cause an accumulation of errors in the subsequent time steps. Thus, when computing the conditional probability $p_{\theta}(y_j|y_{<j},\vec{x})$, Scheduled Sampling \citep{bengio2015scheduled} is often used, a method that alternates between using the model prediction on the previous time step $\hat{y}_{j-1}$ and the target previous word $y_{j-1}$ in order to alleviate the presence of compounding errors.

The word-based Seq2Seq model can capture semantic meaning at a word level and long-term contextual dependencies that help in disambiguation of multiple correction candidates. Figure \ref{fig:example} presents an example of source and target pair of tweets for which context helps in appropriately normalizing the content.

\subsection{Handling unknown words with a secondary character-based encoder-decoder model}

The model operating on words has a limited vocabulary for both source and target. Words that are beyond this vocabulary are represented with a special UNK symbol. For text normalization, where slight variations occur often due to misspellings, keyboard typing errors and intentionally emphasizing terms by elongation of vowels (e.g. ``coooool" or ``yaaaay"), many of the words are unseen during training, resulting in loss of information. Three possible solutions can be used to tackle this problem: a) copying source words, b) rely on models fully trained on character-based information and c) design hybrid models that work both on word and character level.

\begin{figure}[h]
    \centering
    \includegraphics[width=.5\columnwidth]{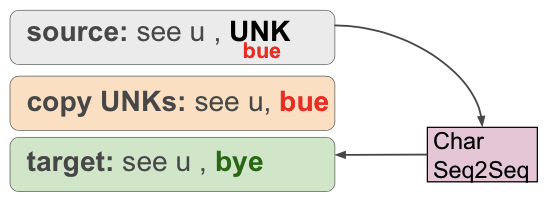}
    \caption{Example of an unseen unnormalized token where copying the source word is insufficient}
    \label{fig:char}
\end{figure}

A naive strategy would be to just copy the source word when it is outside the scope of the vocabulary (see Figure \ref{fig:char}), however many unseen non-standard words will be left intact and thus the coverage of our models will decrease. Another way to handle vocabulary coverage is to pre-process the data and learn a subword representation that allows to generalize to new words. Byte pair encoding (BPE) \citep{sennrich2015neural} learns the segmentation of text into subwords, e.g. ``showed" could be split into ``show" and ``ed" while ``accepting" would be split ``accept" and ``ing". Such pre-processing is model-agnostic, i.e. can be used irrespectively of the chosen Seq2Seq model. However, BPE relies on the cooccurence and order of characters, which in our case is highly noisy. 

Character models overcome the bottleneck of restricted vocabularies and do not require any pre-processing or tokenization but are computationally expensive and also suffer from data sparsity. \citet{chung2016character} provide a detailed analysis regarding the challenges of character-level models. \citet{belinkov2017synthetic} recently showed that character-based models fail to translate noisy text that humans can handle easily. They mention that such models are rarely trained to explicitly handle typos and noise, commonly found in natural language. 

Hybrid models on the other hand, rely primarily on a word-based representation where the meaning is naturally preserved and backtrack to a secondary character-level model to deal with problematic text. Because the character model is trained only in some cases, it requires a large pool of such problematic aligned text. Due to limited training data available for text normalization and the long tail of rare non-standard words, hybrid architectures that train character-level models only for words outside the vocabulary would be insufficient. Thus, for in-vocabulary words we rely solely on the word-level model, while when a word is OOV, we backtrack to a character encoder-decoder that is trained on word pairs rather than longer token sequences, i.e. each pair of source and target words in our training set is processed separately. 

\subsection{Adversarial training for increased robustness to noisy user-generated text}
To improve our model's robustness to noisy text, we incorporate an adversarial training procedure to our character-based secondary model. We augment our data by  creating synthetic adversarial examples of words, i.e. unnormalized and canonical forms. More specifically, for all source-target pairs of tweets, we keep words that remain unchanged. This process creates our source and target vocabularies for the character model. We later on inject multiple types of noise during training, by editing the source part of each word. More specifically, we introduce 6 types of errors that are typically found in user-generated text, by randomly:

\begin{itemize}
    \item[]\textbf{del:} Deleting a character from a word
    \item[]\textbf{swap:}  Swapping the placement of two characters 
    \item[]\textbf{lastchar:}  Elongating the last character $k$ times when the word ends with $\{u, y, s, r, a, o, i\}$, where $k \in \{1,\dots,6\}$ 
    \item[]\textbf{punct:}  Deleting  e.g. ``I'm" $\rightarrow$ ``Im"  or misplacing apostrophes, e.g. ``don't" $\rightarrow$ ``do'nt" 
    \item[]\textbf{keyboard:}  Replacing characters based on their distance on the keyboard, e.g. ``hello" $\rightarrow$ ``jello"
    \item[]\textbf{elong:}  Extending vowel usage $k$ times, where $k$ is a random number
\end{itemize}

\section{Experiments}\label{sec:exp}
In this section we present our experimental setup for assessing the performance of the text normalization model described above. 
We want to test: a) whether a naive replacement of words with their non-standard form would be sufficient for the text normalization task, b) which Seq2Se2 model is the most effective, c) whether BPE and character level models are appropriate for normalizing OOV social media tokens, d) how crucial is context and long-term dependencies for correctly normalizing noisy text and e) whether adversarial training improves robustness of our hybrid architecture.

\subsection{Dataset}
We use the LexNorm dataset from the 2015 ACL-IJCNLP Workshop on Noisy User-generated Text (W-NUT) \citep{baldwin2015shared}. The dataset contains 4,917 tweets with 373 unique non-standard word types, split into 60:40 training/testing ratio. There are 488 non-standard word types that are unseen during training, i.e. not found in the training data. Table \ref{table:stats} lists some statistics of the dataset described in \citep{baldwin2015shared}. Note that apart from mapping a source word to a target word ($1:1$ mapping), there are also words that are mapped to more than one target tokens ($1:N$ mapping), e.g. ``omw" $\rightarrow$ ``on my way".

To reduce vocabulary size, words are lowercased, while mentions were tagged and anonymized with a $\langle mention \rangle$ token. The same anonymization was applied for URLs $\langle url \rangle$ and hashtags $\langle hash\rangle$. At test time, we de-anonymize by looking them up in their source sentences. Additionally, we keep a common vocabulary between source and target text. Each sequence is additionally pre-processed by adding a start $\langle s\rangle$ and end $\langle \setminus s \rangle$ symbol.

\begin{table}
  \centering
  \resizebox{\columnwidth}{!}{
  \begin{tabular}{|l|rrrrrr|r|}
    \hline
    \textbf{Dataset} & \textbf{Tweets} & \textbf{Tokens} &
    \textbf{Noisy} & \textbf{1:1} & \textbf{1:N} & \textbf{N:1} & \textbf{Our Vocab}\\
    \hline
    train  &   2950   &  44385  &  3942  & 2,875 & 1,043 & 10 &  10,084 \\
test  & 1967  &29421   &2776 & 2,024 & 704 & 10 & 7,389 \\
    \hline
    \end{tabular}
    }
      \caption{LexNorm statistics from \citep{baldwin2015shared} and vocabulary statistics after preprocessing}
  \label{table:stats}
\end{table}

\subsection{Baseline models}\label{baselines}
We compare our model (\textbf{HS2S})  with a diverse set of baselines, including two naive dictionary-based approaches: we begin by constructing a lexicon from the training data and correcting only unique mappings (\textbf{Dict1}) or additionally choose randomly when multiple canonical forms are available (\textbf{Dict2}). We also compare with a two-staged strategy that first corrects unique mappings based on the dictionary and secondly utilizes a word-level Seq2Seq model trained to correct only multiple mappings (\textbf{S2SMult}). 

Addditionaly, we include a default attention-based word-level encoder-decoder (\textbf{S2S}) as our baseline for comparison. For this model, OOV words are solely copied directly from the source sequence, thus ``unseen normalizations" are not handled. This baseline should indicate whether targeting unseen normalizations is critical for the model performance. Since character or sub-word level representations can alleviate the problem of limited vocabularies, we also experiment with a character-based model  (\textbf{S2SChar}) and a model that is trained on subwords with BPE tokenization\footnote{We experimented with the original subword implementation (\url{https://github.com/rsennrich/subword-nmt}) as well as a pretrained version \cite{heinzerling2018bpemb} (\url{https://github.com/bheinzerling/bpemb)} that produced better results.} (\textbf{S2SBPE}). 

Finally, we include a model trained on target sequences preprocessed with a special symbol to indicate that the word should be left intact (\textbf{S2SSelf}), e.g. if the source sequence is ``see u soon" and the target sequence is ``see you soon", we replace the target sequence with ``@self you @self".
During prediction, when we generate the normalized target of a source sentence, we replace this special symbol by copying from the source. Ultimately, we seek to find which sequence-to-sequence model is the most effective for the text normalization task.

\subsection{Training Details} 
We keep a shared vocabulary between source and target and also tie the decoder embeddings \citep{press2017using}. We optimized all models with Adam \citep{kingma2014adam} and the gradient is rescaled when the norm exceeds \{5,10\} \citep{pascanu2013difficulty}. Batch size is set to \{32,500\}. All of our models are bi-directional and use attention. To compare performance, we tune each model separately with random search. The best hyper-parameters are summarized in Table \ref{params}. To tune the hyper-parameters we used a $10\%$ random split of the training data and performed random search on the hyper-parameter space. Once the best combination was found, we retrained our system using the full training data set. 

Our adversarial training procedure is guided by an additional hyper-parameter, \textit{noise ratio}, that tunes the number of adversarial instances used. Our best performing model has a noise ratio of $0.1$, i.e., more than $10\%$ of instances used are generated with adversarial training (see Table \ref{table:noiseratio}). In general, we observed that training on large amounts of adversarial instances, e.g. $50\%$ additional instances, results in decreased performance, e.g. when noise ratio is increased to $0.5$, the F1 score is decreased to $82.67\%$ (Table \ref{table:noiseratio}: incorporating large quantities of adversarial instances decreases performance). Furthermore, to reduce the amount of false positives we allow the character-level secondary model to correct only words for confident predictions. If the confidence of the character-level model is low, our architecture copies from the source.

\begin{table}[h]
\centering
\resizebox{\columnwidth}{!}{%
\begin{tabular}{c|c|c|c|c}
 Hyper-parameter & Word-level & secondary Char-level & S2SChar & S2SBPE\\  \hline 
emb.dimension & 100 & 256 & 256 & 100\\ 
neurons/layer & 200 & 500 & 512 & 300 \\ 
layers & 3 & 3 & 3 & 2 \\  
dropout & 0.5 & 0.5 & 0.2 & 0.3 \\ 
learning rate & 0.01 & 0.001 & 0.001 & 0.01 \\
\end{tabular}%
}
\caption{Best performing hyper-parameter settings of our proposed text normalization models}
\label{params}
\end{table}

\begin{table}
  \centering
    \resizebox{\columnwidth}{!}{
  \begin{tabular}{lrrrr}
    \textbf{Model name} & \textbf{Precision} & \textbf{Recall} &
    \textbf{F1} & \textbf{Method highlights}\\
    \hline
    HS2S & 90.66 & 78.14 & 83.94 & Hybrid word-char Seq2Seq\\
    S2S  & 93.39 & 75.75 & 83.65 & Word-level Seq2Seq \\
    \hline
    Dict1 & 96.00 & 52.20 & 67.62  & Dictionary (unique mappings) \\
    Dict2 & 56.27  & 63.57 & 59.70  & Dict1 + Random \\
    S2SMulti & 93.33 & 75.57 & 83.52 & Dict1 + S2S \\
    S2SSelf & 82.74 & 65.50 & 73.11 & @Self for tokens that \\
    & & & & need no normalization\\
    \hline
    \end{tabular}
    }
      \caption{Comparison of our S2S models with word-level baselines.}
  \label{table:submissions1}
\end{table}

\begin{table}
  \centering
    \resizebox{\columnwidth}{!}{
  \begin{tabular}{lrrrr}
    \textbf{Model name} & \textbf{Precision} & \textbf{Recall} &
    \textbf{F1} & \textbf{Method highlights}\\
    \hline
    HS2S & 90.66 & 78.14 & 83.94 & Hybrid word-char Seq2Seq\\
    S2SChar & 67.14 & 70.50 & 68.78 & Character-level Seq2Seq
    \\
    S2SBPE &  20.00  & 52.04 & 28.90 & Word Seq2Seq + BPE  \\
    \hline
    \end{tabular}
    }
      \caption{Comparison of our HS2S model with additional baselines.}
  \label{table:other_baselines}
\end{table}
\subsection{Comparison of Sequence to Sequence models}

\begin{table*}[ht!]
\centering
\begin{minipage}{0.49\textwidth}
\begin{tabular}{|l|l|l|l|}
\hline
Source & Target & Count &Source \\ \hline
u & \{you're, you, u, your\} & 234 & 335 (2) \\
lol & laughing out loud & 197 & 272\\
im & \{i, i'm\} & 153 & 182\\
dont & don't & 57 & 92\\
lmao & laughing my @ss off & 45 & 45\\
n & \{and, in, at, n\} & 40 & 57 (8) \\
omg & oh my god & 34 & 34 \\ \hline

\end{tabular}
\end{minipage}
\begin{minipage}{0.49\textwidth}
\begin{tabular}{|l|l|l|l|}
\hline
Source & Target & Count & Source \\ \hline
2 & \{to, 2\} & 9 & 36 (25)\\ 
ya & \{ya, you, your, yourself\} & 9 & 15 (4) \\
y &  \{y, why\} & 9 & 17 (8) \\
yo &  \{you, your, yo\} & 7 & 12 (1)\\
rt & \{rt, retweet\} & 7 & 602 (582) \\ 
b &  \{b, be, because, by\} & 4 & 20 (9) \\
nah & \{no, nah, not, now\} & 4 & 6 (2) \\
\hline
\end{tabular}
\end{minipage}

\caption{Most frequent correct (\textit{left table}) and incorrect (\textit{right table}) normalizations of our word-level Seq2Seq model. We present how many times a source tweet was (in)correctly normalized (Count column) as well as how many times that term appears in the source-side of the examples (Source). For cases where a token can be normalized to itself, we include how many times that term appears unchanged (information in parentheses)}
  \label{table:examples}
\end{table*}
We compare our novel hybrid model with word-level baseline  Seq2Seq models (Table \ref{table:submissions1}), as well as character-based and BPE-tokenized Seq2Seq models (Table \ref{table:other_baselines}).
In Table \ref{table:submissions1} we see that our dictionary based baselines result in lower performance. \textbf{Dict2} shows that handling ambiguous cases inappropriately results in a dramatic drop in precision. It is therefore necessary for a text normalization model to be able to correctly normalize text in the occurrence of multiple mappings.  \textbf{S2SMulti} is a baseline method that firstly normalizes terms that have a unique mapping, based on source and target tokens found in the training data, and later on utilizes a word-level Seq2Seq model that is trained to correct one-to-many mappings (an unnormalized token that can be transformed into several standard words, e.g. ``ur" $\rightarrow$ \{``you are", ``your"\}). We can see that this method has a better performance, which validates our hypothesis that a naive word replacement would not suffice, however the end-to-end  Seq2Seq model \textbf{(S2S)} performs better than \textbf{S2SMulti}, i.e. Seq2Seq models can handle both unique and multiple mappings seamlessly without any additional feature engineering effort apart from tokenization. 

Our character-level \textbf{S2SChar} model's performance is slightly above the dictionary baseline \textbf{Dict1} which suggests that characters do not contain enough semantics to appropriately disambiguate between terms. Our results align with relevant literature \citep{belinkov2017synthetic} that emphasizes on the noise sensitivity of character-based Seq2Seq models. We should also note that character-level Seq2Seq models take longer to train. Our best performing word Seq2Seq model took 22 minutes to train while our top scored character model took 3 hours, for the same number of epochs. The secondary character model of our hybrid architecture, which is trained on pairs of words, took 42 minutes for the same number of epochs.  

Despite extensively tuning the hyper-parameters and experimenting with two subword tokenization tools, we were unable to train a good performing model on subword units (\textbf{S2SBPE}) successfully. S2SBPE has surprisingly very low performance, the poorest of all models. As BPE relies on co-occurrence of characters to extract frequent subword patterns, one would expect that it would not be able to capture useful information due to the high percentage of noise. This emphasizes the importance of developing models robust to informal text, that can learn from noisy input ``on the fly".

\subsection{Error analysis}
We perform an extensive error analysis. First, we check the model output, particularly in which cases our model fails. Table \ref{table:examples} presents the most frequent normalizations that our model performed correctly and the most frequent cases that were missed, as well as the frequency of the source terms. Note that we also keep track of how many times a term remains unchanged, specifically for cases where that term has multiple mappings available  (Table \ref{table:examples}, information found in parentheses). We can see that our model can handle multiple mappings when those are adequately represented in the training data. Most of the incorrect normalizations appear less than 50 times in our data (very infrequent) or are ambiguously normalized in some examples and left intact in other examples, e.g. ``rt" and ``2" remain unchanged a few times, and these are the cases that the model missed.  These types of errors can be handled by adding more training data. Furthermore, we should note that our model can also be adjusted to work with distantly supervised data or with ensemble methods.

Comparing with different representations that can handle unknown words, our character level model performs edits that might result in removing words from the text or are close to the gold-truth canonical form but not entirely correct (see Table \ref{table:sent_mod}). In contrast to our hybrid model that can preserve contextual and word-level semantic information, our BPE model's poor performance results in editing text that needs no correction.

Futhermore, we perform error analysis on the secondary character-level model that is trained on synthetic adversarial examples of word pairs. In total, our model correctly normalizes $16.22\%$ of unseen source words. In Table \ref{table:confidence} we present OOV terms that the model can correctly normalize. Most frequent errors that are correctly normalized are elongating the last characters, deleting last character [$g$] in gerund, swapping or replacing characters and typos. There are also several typos that the model was not able to correct, such as missing whitespaces or editing words unnecessarily. 

\begin{table}
  \centering
    \resizebox{\columnwidth}{!}{
  \begin{tabular}{lcccccc}
    \begin{tabular}[c]{@{}c@{}}\textbf{Noise} \\ \textbf{Ratio} \end{tabular} & \begin{tabular}[c]{@{}c@{}}\textbf{Total}\\ \textbf{examples}\end{tabular} & \begin{tabular}[c]{@{}c@{}}\textbf{Noise-injected}\\ \textbf{examples}\end{tabular} & \textbf{Precision} & \textbf{Recall} & \textbf{F1} \\
    \hline
    0.1 & 34,875 &  5,739 & 90.66 & 78.14 & 83.94 \\
    0.2 & 37,659 &   8,523 &  89.92 & 78.25 & 83.68 \\
    0.3 & 40,489 &  11,353 &  88.61 & 78.11 & 83.03 \\
    0.4 & 43,191 &  14,055 & 89.25 & 78.25 & 83.39 \\
    0.5 & 45,921 &  16,155 & 87.33 & 78.47 & 82.67 \\
    0.6 & 48,625 &  19,489 & 86.21 & 79.05 & 82.47 \\
    0.7 & 51,380 &  22,244 & 84.89 & 78.83 & 81.75 \\
    0.8 & 54,034 &  24,898 & 84.37 & 79.05 & 81.62 \\
    0.9 & 56,829 &  27,693 & 83.94 & 78.98 & 81.38 \\
    \hline
    \end{tabular}
    }
      \caption{Varying the amount of adversarial examples}
  \label{table:noiseratio}
\end{table}

\begin{table}[h]
  \centering
  \resizebox{\columnwidth}{!}{%
  \begin{tabular}{ll}
  \textbf{Source:} & cmon familia dont mess this up please  \\
   \textbf{Target:} &come on familia don't mess this up please   \\
    \textbf{HS2S/S2Multi:} & {\color{blue}come} on familia don't mess this up please \\ 
    \textbf{S2SSelf:} & {\color{red}cmon} familia don't mess this up please \\
    \textbf{S2Char:}  & {\color{red}comon} familia don't mess this up please \\
    \textbf{S2SBPE:} & {\color{red}cmon} familia don't just ess this up \\
    \hline
    \textbf{Source} & ... i'm not gon diss you on the internet cause ...\\
    \textbf{Target:} & ... i'm not gonna {\color{blue}disrespect} you on the internet because ...\\
    \textbf{HS2S/S2Multi:} & ... i'm not gonna {\color{blue}disrespect} you on the internet because ... \\
      \textbf{S2Char:} & ... i'm not gonna {\color{red}thiss} you on the internet because ...\\
    \textbf{S2SBPE:} & ... i'm not gonna {\color{red}be} you on the internet because ... \\
    \end{tabular}
    }
      \caption{Examples where our model surpasses architectures that rely on lower-level representation of text.}
  \label{table:sent_mod}
\end{table}

\begin{table*}[h]
\centering
\begin{minipage}{0.43\textwidth}
\begin{tabular}{lrrrr}
    \textbf{Source} & \textbf{Prediction} & \textbf{Confidence $(\%)$} \\
    \hline
   perfomance	& {\color{blue} performance}& 83.90 \\
   considerin & {\color{blue} considering} & 78.94 \\
   birthdayyyyy & {\color{blue} birthday} & 74.30 \\
   brothas & {\color{blue} brother} & 72.53 \\
   pepole & {\color{blue} people} & 72.02 \\
   tomorroww & {\color{blue} tomorrow} & 69.03 \\
   yesssssss & {\color{blue} yes} & 68.35 \\
   chillll & {\color{blue} chill} & 60.45 \\
   iight & {\color{blue} alright} & 59.35  \\
    \end{tabular}
    \end{minipage}
\begin{minipage}{0.43\textwidth}
  \begin{tabular}{lrrrr}
        \textbf{Source} & \textbf{Prediction} & \textbf{Confidence $(\%)$} \\
    \hline
   pantsareneveranoption	& {\color{red} pantsarentien}& 83.29 \\
   judgemental	& {\color{red} judgmmental}& 79.11 \\
   kissimmee	& {\color{red} kissimme}& 74.08 \\
   bsidez	& {\color{red} baidez}& 67.67 \\
   happppy	& {\color{red} happy}& 61.67 \\
   coldplay	& {\color{red} coldolay}& 59.43 \\
   knob	& {\color{red} know} & 58.12 \\
   donuts	& {\color{red} doughs} & 57.78 \\
   becos	& {\color{red} becouse}& 55.72 \\
   \end{tabular}
\end{minipage}
      \caption{OOV words that our secondary character model has normalized correctly (blue) or incorrectly (red)}
  \label{table:confidence}
\end{table*}

\subsubsection{How contextual information affects performance}
\begin{table}[h]
  \centering
  \resizebox{\columnwidth}{!}{%
  \begin{tabular}{ll}
  \textbf{Source:} & think tht took everything off ma mind for tha night \\
   \textbf{Target:} & think that took everything off my mind for the night  \\
    \textbf{HS2S:} (80\%) & think {\color{blue}that} took everything off {\color{red}ma} mind for {\color{blue}the} night \\
    \textbf{S2SSelf:} (50\%) & think {\color{blue}that} took everything off {\color{red}ma} mind for {\color{red}the} {\color{red}tha} night \\
    \hline
    \textbf{Source:} &  death penalty would b d verdict @general$\_$marley murder will b d case  ... \\
    \textbf{Target:} & death penalty would be the verdict @general$\_$marley murder will be the case  ...\\
    \textbf{HS2S:} (88.8\%) &  death penalty would {\color{blue}be the} verdict @general\_marley murder will {\color{red}b} {\color{blue}the} case  ... \\
    \textbf{S2SSelf:} (0\%) & death penalty would {\color{red}b d} verdict @general$\_$marley murder will {\color{red}b d} case  ...\\
    \end{tabular}
    }
      \caption{Comparing HS2S with S2SSelf shows context is crucial for correct normalization, especially for short tokens.}
  \label{table:self}
\end{table}

One of our main questions is how crucial contextual information is for text normalization. We test the importance of context by performing two experiments. For our first experiment, we perform the following preprocessing of the training data: we constrain the target-side of each example by replacing each word that remains unchanged (no normalization needed) with a special \textit{@self} symbol. With this representation all tokens that need normalization are preserved and the model would learn which words remain the same. We call this model \textbf{S2SSelf}. We notice that this representation lowers performance due to loss of contextual information on the target side (see Table \ref{table:submissions1}). In Table \ref{table:self} we present examples of our hybrid model's predictions and compare with the predictions of S2SSelf. In most cases we observe that our model relies on context to normalize short tokens, while S2SSelf fails to correct such terms.

\begin{figure}[h]
    \centering
    \includegraphics[width=\columnwidth]{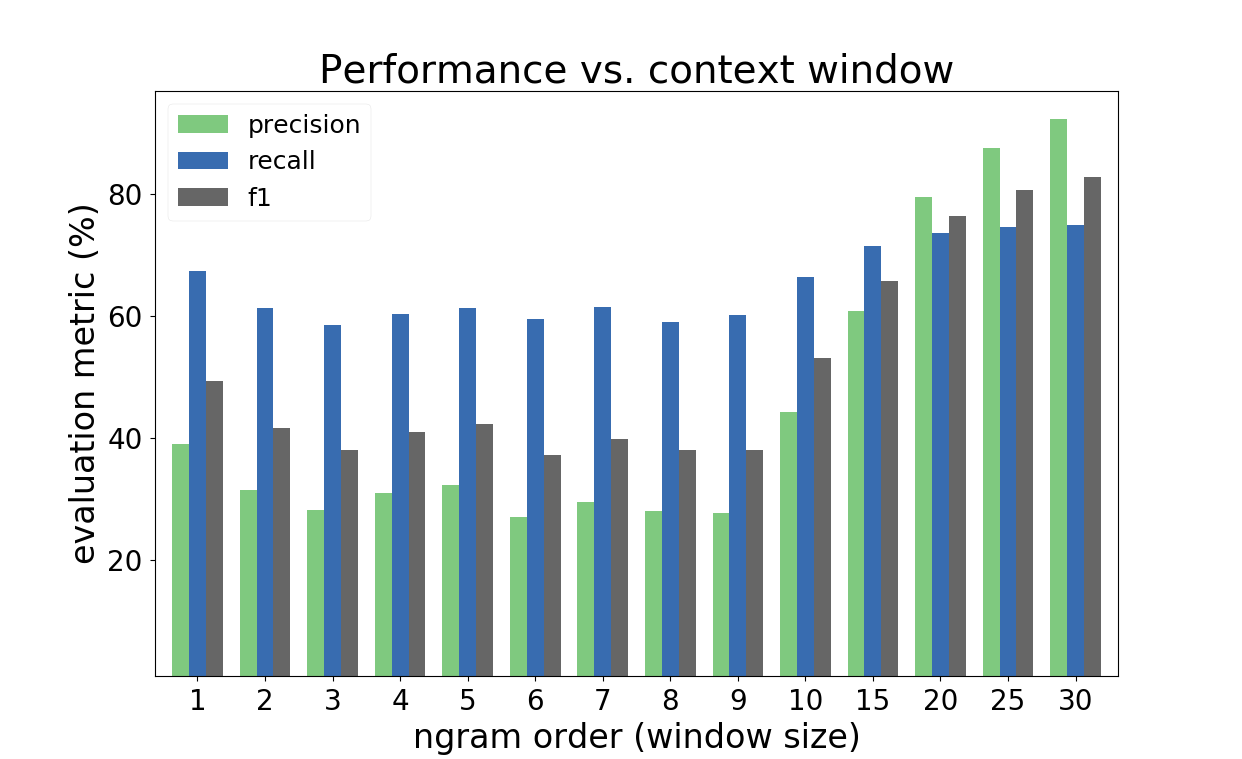}
    \caption{Varying the ngram-wise split of sequences to check how context affects performance of text normalization.}
    \label{fig:ngram}
\end{figure}

Moreover, we create ngram representations of our data by splitting the tweets: for example a unigram model is trained on word pairs solely and ignores context when normalizing, as it edits each word separately and similarly a bigram model is trained on phrases that contain two words. We continue with higher-order ngrams, train Seq2Seq models on such ngram-based split of text and analyze the importance of contextual information. By gradually varying the context window, while keeping the rest of the hyper-parameters stable, we can analyze how it affects the performance. In Figure \ref{fig:ngram} we can see that recall remains fairly unchanged, while precision increases as the context window grows larger, i.e. train on higher-order ngrams and thus incorporating more context. Overall, we can observe that F1 measure gets better with additional contextual information.  

\subsection{Comparison with related work}
Finally, we present our comparison with related work on Table \ref{table:submissions2}. We see that all previous Deep Learning approaches are close to 82\% F1 score. Due to the nature of our hybrid model, we were able to achieve the best performance so far among neural models in related work. In general, we observe comparable performance with state-of-the-art methods that are constrained on utilizing additional resources\footnote{MoNoise \citep{van2017monoise} leverages large collections of Twitter and Wikipedia data.}. We compare the incorrect normalizations that our Seq2Seq model and MoNoise - the best performing method - produce. Both systems appear to have similar results in terms of most frequent incorrect normalizations (Table \ref{table:jinboth}). In many cases our hybrid Seq2Seq model leaves intact terms that are ambiguous in terms of whether they should be normalized or not, while \citep{van2017monoise} normalize words more often (Table \ref{table:jindiff}), in some cases incorrectly.  

\hide{Additionally, we have implemented the forest approach\footnote{Our implementation achieved $83.04\%$ F1 score, slightly less than \citet{jin2015ncsu} due to insufficient information on how splitting phrases was resolved.} (\textbf{Our\_RF} in Table \ref{table:submissions2})  and compared the incorrect normalizations that our Seq2Seq model and the random forest produce. Both systems appear to have similar results in terms of most frequent incorrect normalizations (Table \ref{table:jin}).}

\begin{table}[h]
  \centering
  \resizebox{\columnwidth}{!}{
  \begin{tabular}{lrrr}
    \textbf{Model} & \textbf{Precision} & \textbf{Recall} &
    \textbf{F1} \\ \hline
   Hybrid Seq2Seq (HS2S) & 90.66 & 78.14 & 83.94 \\ 
   Random Forest \citep{jin2015ncsu}& 90.61  & 78.65  & 84.21 \\
   Lexicon +LSTM \citep{min2015ncsu_sas_wookhee}& 91.36  & 73.98  & 81.75  \\
   ANN \citep{leeman2015ncsu_sas_sam}& 90.12  & 74.37  & 81.49  \\
    MoNoise{*} \citep{van2017monoise} & 93.53 & 80.26 & 86.39 \\ \hline
    \end{tabular}
    }
      \caption{Comparison of our hybrid Seq2Seq model with related work on Text Normalization. {*}{In contrast with the rest of the presented related work, Monoise leverages additional textual resources.}}
  \label{table:submissions2}
\end{table}

\begin{table}
  \centering
  \resizebox{\columnwidth}{!}{%
  \begin{tabular}{ll}
  \textbf{Source:} & @ifumi0819 i see , u can comeee\\
   \textbf{Target:} & @ifumi0819 i see , you can come  \\
    \textbf{HS2S:} & @ifumi0819 i see , you can {\color{red} comeee } \\
    \textbf{Our\_RF:} & @ifumi0819 i see , you can {\color{red} comes } \\
    \hline
    \textbf{Source:} & startin to get into this type of musik @vinnyvitale\\
    \textbf{Target:} & starting to get into this type of music @vinnyvitale\\
    \textbf{HS2S:} & {\color{blue} starting} to get into this type of {\color{red} musik } @vinnyvitale \\
    \textbf{Our\_RF:} & {\color{red}  startin} to get into this type of {\color{red} musik } @vinnyvitale \\ \hline
    \textbf{Source:} & \#youarebeautiful allly this hashtag should be for you im ugly\\
    \textbf{Target:} & \#youarebeautiful allly this hashtag should be for you i'm ugly\\
    \textbf{HS2S:} & \#youarebeautiful allly this hashtag should be for you {\color{blue} i'm} ugly \\
    \textbf{Our\_RF:} &  \#youarebeautiful {\color{red} alli} this hashtag should be for you {\color{blue} i'm} ugly  \\
    \end{tabular}
    }
      \caption{Examples of corrent and incorrect normalizations of our model (HS2S) and \citet{jin2015ncsu} (our implementation)}
  \label{table:jin_wrong}
\end{table}

Interestingly, there are examples that, despite the lack of correct target annotation, our model normalizes tokens correctly (see Table \ref{table:sente}). More specifically, there are several tokens that were not normalized but in an ideal scenario should be, e.g. many punctuation errors that were not normalized or abbreviated tokens that were not converted into their standard form. As a result, despite correctly transforming text, our system gets penalized for such cases.

\begin{table}[h]
  \centering
  \resizebox{\columnwidth}{!}{%
  \begin{tabular}{ll}
  \textbf{Source:} & rt @foxtramedia tony parker sportscenter convo ... \\
   \textbf{Target:} & rt @foxtramedia tony parker sportscenter {\color{red}convo} ... \\
    \textbf{Prediction:} & rt @foxtramedia : tony parker sportscenter \textit{\color{red}{conversation}} ... \\
    \hline
    \textbf{Source:} & ... but looking back defo do now, haha ! \\
    \textbf{Target:} &  ... but looking back {\color{red}defo} do now, haha ! \\
    \textbf{Prediction:} & ... but looking back \textit{\color{red}{definitely}} do now, haha ! \\
    \end{tabular}
    }
      \caption{Examples where our model performs correct normalization but during the annotation process, some tokens remained unnormalized in the target sequence, which results in lower performance in evaluation.}
  \label{table:sente}
\end{table}

\begin{table}[h]
\centering
\resizebox{\columnwidth}{!}{
\begin{tabular}{|c|c|c|c|c|c|} \hline
\textbf{Source} & \textbf{Target} & \textbf{HS2S (ours)} & \textbf{MoNoise}\\  \hline 
youd & would & youd &   you'd \\  
ya & your & you & ya \\  
yknow & you know & yknow & know \\
werent & weren't & werent & were \\
tiz & this & tizket & tiz \\
swthat & shout out & switht & swthat \\
shite & shitty & shit & shite \\
rts & retweets & rts & rts \\
pleeze & please & pleaze & pleeze \\
nah & now & no & nah \\
judgemental & judgmental & judgmmental &  judgemental \\
championssssss & champions & championsssss & championssssss \\ \hline
\end{tabular}%
}
 \caption{Examples that both our hybrid (HS2S) model and MoNoise \citep{van2017monoise} incorrectly normalized.}
  \label{table:jinboth}
\end{table}

\begin{table}[h]
\centering
\resizebox{0.9\columnwidth}{!}{
\begin{tabular}{|c|c|c|c|} \hline
 & \textbf{Source} & \textbf{Target} & \textbf{Prediction} \\  \hline
\multirow{9}{*}{\begin{tabular}[c]{@{}l@{}}\textbf{HS2S} \\ (ours)\end{tabular}} 
& 2night & tonight & 2night \\
& aboul & about & aboul \\
& asap & as soon as possible & asap \\
& bermudez & bermudez & bermudes \\
& bfor & before & boor \\
& cruz & cruz & crus \\
& outta & outta & out of \\
& pppp & people & pppp \\ \hline
\multirow{9}{*}{\begin{tabular}[c]{@{}l@{}}\textbf{MoNoise} \\\end{tabular}}  
& wildin & wilding & wildin \\ 
& weeknd & weekend & weeknd\\
& tix & tickets & ticket\\
& da & the & da \\
& rip & rest in peace & rip \\
& probs & problems & probably \\
& gf & girlfriend & gf \\
& broo & brother & bro \\ \hline
\end{tabular}%
}
 \caption{Examples of incorrect normalizations that are distinct between our hybrid (HS2S) model and and MoNoise \citep{van2017monoise}.}
  \label{table:jindiff}
\end{table}

\section{Conclusions and future work}\label{sec:conclusion}
Text normalization is an important preprocessing part that helps users understand online content and increases performance of off-the-shelve pretrained NLP tools. However, due to the inherent constraints of existing feature engineering methods used, existing work cannot capture longer contextual information and is limited to handling specific types of normalization corrections. Neural Seq2Seq models can naturally correct complex normalization errors by learning edits on large pools of text data. Additionally, improving robustness of Seq2Seq models on real-word noisy text data is a crucial problem that remains fairly unexplored.  To this end we have introduced a novel hybrid neural model for social media text normalization that utilizes a word-based encoder-decoder architecture for IV tokens and a character-level sequence-to-sequence model to handle problematic OOV cases. Our character-based component is trained on adversarial examples of word pairs. Experimental results show that our hybrid architecture improves robustness to noisy user-generated text and shows superior performance, when compared with open vocabulary models. Without relying on any external sources of additional data, we built a system that improves the performance of neural models of text normalization and produces results comparable with other models found in the recent related literature. Our system can be deployed as a preprocessor for various NLP applications and off-the-shelve tools to improve their performance on social media text.  

We plan to apply the approach to more languages and compare our adversarial training to other methods, e.g. perturbations applied directly to the embedding space instead of the input. While normalizing informal text, it is worth to consider whether the meaning of a noisy version remains the same, for example the extended usage of vowels (``yaaaaaaay") indicates emphasis while capitalization represents raising the tone. We leave the analysis of the trade-off between retaining such information and normalizing noisy text as future work.

\subsubsection{Acknowledgments.}
The authors would like to thank the anonymous reviewers for their helpful comments. This material is based upon work supported by the National Science Foundation under Grant No. 1801652.


\bibliographystyle{aaai}
\bibliography{refs.bib}

\end{document}